\documentclass{article}
\usepackage{ijcai24}

\usepackage{fancyhdr}
\usepackage{times}
\usepackage{soul}
\usepackage{url}
\usepackage[hidelinks]{hyperref}
\usepackage[utf8]{inputenc}
\usepackage[small]{caption}
\usepackage{graphicx}
\usepackage{amsmath}
\usepackage{amsthm}
\usepackage{booktabs}
\usepackage{algorithm}
\usepackage{algorithmic}
\usepackage[switch]{lineno}

\usepackage{natbib}
\usepackage{mathptmx}
\usepackage{booktabs}
\usepackage{array}
\usepackage{mathtools}
\usepackage{amssymb}
\usepackage{xcolor}

\title{It Ain't That Bad: Understanding the Mysterious Performance Drop in OOD Generalization for Generative Transformer Models}

\author{
Xingcheng Xu$^{1}$\and
Zihao Pan$^{2}$\and 
Haipeng Zhang$^{2}$\thanks{Corresponding authors.}\And
Yanqing Yang$^{1,3*}$\\
\affiliations
$^1$Shanghai Artificial Intelligence Laboratory\\
$^2$ShanghaiTech University\\
$^3$Fudan University\\
\emails
xingcheng.xu18@gmail.com,
\{panzh,zhanghp\}@shanghaitech.edu.cn,
yanqingyang@fudan.edu.cn
}

\begin{document}

\maketitle
\thispagestyle{fancy} 

\begin{abstract}
Large language models (LLMs) have achieved remarkable proficiency on solving diverse problems. However, their generalization ability is not always satisfying and the generalization problem is common for generative transformer models in general. Researchers take basic mathematical tasks like $n$-digit addition or multiplication as important perspectives for investigating their generalization behaviors. It is observed that when training models on $n$-digit operations (e.g., additions) in which both input operands are $n$-digit in length, models generalize successfully on unseen $n$-digit inputs (in-distribution (ID) generalization), but fail miserably on longer, unseen cases (out-of-distribution (OOD) generalization). We bring this unexplained performance drop into attention and ask whether there is systematic OOD generalization. Towards understanding LLMs, we train various smaller language models which may share the same underlying mechanism. We discover that the strong ID generalization stems from structured representations, while behind the unsatisfying OOD performance, the models still exhibit clear learned algebraic structures. Specifically, these models map unseen OOD inputs to outputs with learned equivalence relations in the ID domain, which we call the \textit{equivalence generalization}. These findings deepen our knowledge regarding the generalizability of generative models including LLMs, and provide insights into potential avenues for improvement.
\end{abstract}

\section{Introduction\label{sec:Introduction}}

Large language models (LLMs) such as ChatGPT~\citep{ouyang2022training}, GPT-4~\citep{OpenAI2023GPT4TR}, Claude~\citep{Anthropic2023Claude}, PaLM~\citep{chowdhery2023palm}, Llama~\citep{touvron2023llama,touvron2023llama2} have exhibited remarkable advancements across
diverse domains, prominently in natural language processing (NLP). The LLMs have demonstrated exceptional versatility, showcasing profound efficacy in tackling a myriad of tasks, ranging
from natural language challenges to code translation, mathematical
reasoning, and more~\citep{bubeck2023sparks, trummer2022codexdb, zong2023solving}. 
Although these accomplishments are undoubtedly impressive, the generalization ability of LLMs and generative transformer models in general is not fully understood and not always satisfactory in issues such as natural language understanding~\citep{bender2021dangers}, and mathematical reasoning~\citep{anil2022exploring}. 

Given the complexity of natural language tasks and the black-box nature of these models, researchers view basic mathematical tasks such as $n$-digit addition or multiplication as valuable avenues for gaining insights into their generalization behaviors~\citep{lee2023teaching, anil2022exploring}. Among them, many have observed an interesting yet mysterious phenomenon when training on $n$-digit operations~\citep{brown2020language,anil2022exploring,jelassi2023length}. In cases where both input operands are $n$-digit long, the models demonstrate excellent generalization on unseen $n$-digit inputs. However, they unexpectedly and miserably struggle when faced with longer, unseen cases (inputs with more than $n$ digits). For instance, when trained with operations like $349+705=1054$, the model would perform well on unseen input $350+705$. But when the inputs are $1349+2705$ which are longer in digits, the model gives a wrong answer. This creates a clear distinction between the former, known as \textit{in-distribution (ID) generalization}, and the latter, termed \textit{out-of-distribution (OOD) generalization}.

Seeking to bridge this generalization gap, scholars have undertaken various efforts to enhance OOD generalization. The techniques employed in this pursuit encompass a diverse spectrum, including modifying position embeddings~\citep{jelassi2023length} and attention mechanisms~\citep{dubois2019location}, fine-tuning using extended data samples, prompting and Scratchpad~\citep{anil2022exploring}, priming through selective longer-length data~\citep{jelassi2023length}, and even utilizing chain-of-thought (CoT) style data~\citep{lee2023teaching}. 

In spite of these different techniques, there is still a lack of understanding regarding the underlying mechanism. The proposed solutions may therefore have questionable robustness and become vulnerable to circumstance changes~\citep{jelassi2023length}. Considering the evident and notably poor OOD performance, it is natural to ask whether it stems solely from random errors or if there is anything informative learned by these models.

In this paper, we bring the mystery into attention and seek from the mechanistic perspective~\citep{nanda2022mechanistic, zhong2023clock} in model interpretability. This avenue of study offers a macroscopic understanding of how neural networks work and has helped identify and interpret significant phenomena such as ``grokking", also known as delayed generalization where models exhibit improved generalization long after over-fitting their training set~\citep{liu2022towards}.

When conducting experiments, it is intuitive to test models with ID samples as well as OOD ones to make comparisons. However, it is not feasible if we use well-known LLMs such as GPT-4 or Llama, since we do not know the exact data that they are trained on and therefore cannot distinguish between ID and OOD samples. On the other hand, training LLMs is inevitably very expensive~\citep{brown2020language}. Nonetheless, the study by \cite{anil2022exploring} shows that when the model scale increases, the model ability to generalize across different task lengths does not improve. This suggests that the underlying mechanism may be irrelevant to model scale and all generative models may share the same mechanism. Inspired by this and just like many other studies~\citep{lee2023teaching,jelassi2023length}, we dig smaller models for insights that could apply to LLMs. Besides, we further increase the model scales to examine the consistency in the range of scales that we can reach.

Tasks such as $n$-digit (modular) addition and multiplication are tools for investigating issues including length generalization \citep{anil2022exploring} and ``grokking'' \citep{liu2022towards}. Albeit simple, they offer clearer, more controlled conditions, which can lead to more reliable observations and interpretations. In this paper, through 
training a set of small generative language models, including NanoGPT and MinGPT~\citep{karpathy2022mingpt}, on $n$-digit addition and multiplication tasks, we have made an intriguing discovery. We find that the strong ID generalization stems from structured representations, while the models have learned a clear algebraic structure behind the unsatisfying OOD performance. Specifically, these models map unseen OOD inputs to outputs with equivalence relations in the ID domain, which we call the phenomenon as \textit{equivalence generalization}. The representation learning process plays a crucial role in facilitating both ID and OOD generalization observed in these models. Initially, the representations are random. But as training progresses, the structure of the learned representations becomes increasingly refined,equivalence generalization eventually allowing the models to accurately encode every input in the ID domain. Concurrently, these structured representations are continuously extended to map the unseen OOD domain. However, this extension does not occur as ideally anticipated, resulting in the poor OOD performance. Thus, the representation learning enables powerful ID generalization, but the continuous extrapolation of these representations to OOD inputs gives rise to systematic, rather than random, errors. The mechanistic insights from the discovered patterns also highlight the potential of these models to make use of the information for better generalization.

As a note, we perform several robustness studies in this work, such as changing the encoding method and varying the training data scheme. We find that the equivalence generalization phenomenon is robust. In addition, we conduct a detailed examination of the results across different model scales. Notably, our results remain consistent as the model scales increase, which strengthens our confidence that these results might be extended to LLMs. 

Our main contributions are as follows: 
\begin{itemize}

\item \textbf{Showcasing the power of mechanistic empirical evaluation for LLM generalization}:
We train small generative language models (e.g., NanoGPT, MinGPT) on arithmetic tasks to directly investigate ID vs. OOD generalization, rather than resorting to workarounds. As a result, our approach provides macroscopic insights. To facilitate relevant research, we opensource our code\footnote{The code is available at \url{https://github.com/xingchengxu/ExploreGPT}}.

\item \textbf{Discovering learned structure for OOD generalization}: 
The discernible algebraic structure and the equivalence generalization would hopefully guide robust essential solutions for strong OOD generalization.

\item \textbf{Understanding the role of representations in generalization}: We show that representation learning enables strong ID performance, while unanticipated extension of representations to OOD inputs leads to systematic errors.

\end{itemize}

\section{Related Work\label{sec:Related}}

\subsection{Generalization of Language Models in Arithmetic}
Various studies have examined the performance of Transformer-based language models in tasks involving arithmetic operations. 
\citet{brown2020language} investigated the ability of GPT-3 to perform basic arithmetic operations without task-specific training.
\citet{nogueira2021investigating} explored the limitations of transformers in handling simple arithmetic operations.
Subsequent studies have further explored the generalization capabilities of language models in arithmetic tasks. \citet{qian2022limitations} discovered that language models exhibit poor OOD generalization, and traditional methods such as explicit positional markers and fine-grained computation steps do not effectively address this issue.
To enhance the generalization ability of the model, certain studies have approached the issue starting from a microscopic perspective. For instance, \citet{jelassi2023length} replaced absolute position embeddings with relative position embeddings. Additionally, \citet{dubois2019location} suggested that utilizing a location-based attention mechanism proves effective in the Lookup Table task. 
Other research has focused on the intermediate learning process of the model.
\citet{anil2022exploring} observed that requesting the model to generate intermediate arithmetic steps before providing the final output can improve generalization. \citet{jelassi2023length} arrived at similar conclusions by decomposing the arithmetic pipeline and improving generalization in five-digit addition tasks.
In contrast, \citet{lee2023teaching} presented a different perspective, emphasizing the importance of high-quality, instructive data that can quickly elicit arithmetic capabilities.

While previous studies have primarily focused on evaluating or improving the generalization capabilities of language models, our work has a different objective, we aim to uncover the underlying mechanisms that govern generalization. This explanatory goal, which seeks to understand the foundations of generalization, has not been explicitly addressed in prior research.

\subsection{Mechanistic Interpretability}
Neural network interpretation has seen numerous studies focusing on various types of models, including deep neural networks (DNNs)~\citep{nam2020relative,barbiero2022entropy}, convolutional neural networks (CNNs)~\citep{yuan2019interpreting, akhtar2020interpretation}, and graph neural networks (GNNs)~\citep{yuan2020xgnn, xuanyuan2023global}. 
These works demonstrate diverse microscopic interpretation techniques tailored to different architectures. 
From a macroscopic perspective, \citet{liu2022towards} tackle delayed generalization or ``grokking" using addition and modular addition tasks. They provide intuitive explanations using effective theories and phase diagrams. Similarly, \citet{zhong2023clock} use modular addition to mechanistically explain algorithm discovery in neural networks. Our work contributes to this growing field of mechanistic interpretability by offering a macroscopic explanation specifically for generative Transformer models.

\section{Preliminary and Experimental Setup\label{sec:Methods}}

\subsection{Model Details}
We employ the model framework of GPT, a Transformer with a decoder-only
architecture comprising multiple layers and multi-head attentions.
We train several small-scale models, namely NanoGPT and MinGPT~\cite{karpathy2022mingpt}, from random initialization using
character-level tokenization and the conventional next-token prediction
objective. The training is conducted on basic mathematical operations,
specifically addition and multiplication of integers. 
Detailed hyperparameters are shown in Table~\ref{table:hyperparameters}. 

\begin{table}[htb]
\small
\begin{centering}
\begin{tabular}{>{\centering}m{2.5cm}>{\centering}m{2.1cm}>{\centering}m{2.1cm}}
\toprule 
\centering{}\textbf{Hyperparameter} & \centering{}\textbf{Addition} & \textbf{Multiplication}\tabularnewline
\midrule
num layer & 3 & 6\tabularnewline
num head & 3 & 6\tabularnewline
dim embd  & 48 & 192\tabularnewline
vocab size & 10  & 10\tabularnewline
context window & 15 & 19\tabularnewline
dropout prob & 0.1 & 0.1\tabularnewline
optimizer & AdamW & AdamW\tabularnewline
learning rate & 0.0005 & 0.0005\tabularnewline
betas & (0.9, 0.95) & (0.9, 0.95)\tabularnewline
weight decay & 0.1 & 0.1\tabularnewline
grad norm clip & 1.0 & 1.0\tabularnewline
\bottomrule
\end{tabular}
\par\end{centering}
\caption{Hyperparameter Information}
\label{table:hyperparameters}
\end{table}

\subsection{Dataset}
The dataset is structured as a concatenation of operand pairs in a
natural order, with the reversed order of the operation results. This
format, demonstrated to be more conducive for learning in next-token
prediction models~\cite{lee2023teaching}, offers a more approachable learning
process. For instance, consider the 3-digit addition $a+b=c$, represented
as ``$a_{2}a_{1}a_{0}+b_{2}b_{1}b_{0}=c_{3}c_{2}c_{1}c_{0}$" in the
standard format. By reversing the output order of ``$c$", we obtain
the reversed data format ``$a_{2}a_{1}a_{0}+b_{2}b_{1}b_{0}=c_{0}c_{1}c_{2}c_{3}$".
As we train addition and multiplication models as distinct entities,
we omit both the operation symbols, i.e., $+$ and $ \times$, and the equal sign, i.e., $=$, from the dataset.
Subsequently, the data undergoes character-level tokenization, resulting
in a vocabulary size of 10, corresponding to digits from 0 to 9. When
the context window surpasses the requisite size for a 3-digit addition,
we pad zeros before numbers ``$a$", ``$b$", and ``$c$".  For instance, in the case of 3-digit addition with a context window of 15, the addition expression ``$349 + 705 = 1054$'' will be encoded as ``$0034900705450100$''.

The dataset is partitioned into three distinct subsets: the training
set $\mathcal{D}_{1}$, randomly sampled from $n$-digit operations;
the test set $\mathcal{D}_{2}$, also drawn from $n$-digit operations
but intentionally devoid of any overlap with the training set (termed
as the ID test set); and an additional test set $\mathcal{D}_{3}$,
sampled from $m$-digit operations with $m>n$, where the value at
positions greater than $n$ is non-zero (referred to as the OOD test set).

In the experiments, we set $n=3$ and $m=5$ for both addition and multiplication operations.  Subsequently, from each of the datasets $\mathcal{D}_{1}$, $\mathcal{D}_{2}$, and $\mathcal{D}_{3}$, we select 10,000 data points as the training set for addition and 50,000 for multiplication. We sample 10,000 for the ID test set and OOD test set, respectively for both operations.

\subsection{ID and OOD Domains}

The data space is compartmentalized into three non-overlapping
regions: $\ensuremath{\mathcal{D}_{1}},\ensuremath{\mathcal{D}_{2}},$
and $\mathcal{D}_{3}$. The union of $\mathcal{D}_{1}$ and $\mathcal{D}_{2}$
constitutes an ID domain, whereas $\mathcal{D}_{3}$
represents an OOD domain. The models learn a function
\[
f:\mathcal{D}_{1}\cup\mathcal{D}_{2}\cup\mathcal{D}_{3}\rightarrow\mathcal{S},
\]
where $\mathcal{S}$ could be the output operation result space $\mathbb{N}$, the output probability space, or the learned representation space.

Since the model is exclusively trained on the $\mathcal{D}_{1}$ space,
the acquired knowledge concerning $\mathcal{D}_{2}$ and $\mathcal{D}_{3}$
is an extension of$\mathcal{D}_{1}$, albeit with an unclear underlying
structure. This constitutes the core aspect we seek to understand.

\subsection{Equivalence Classes}
When training for addition and multiplication on $n$-digit operations,
we have identified a discernible algebraic structure. This is encapsulated
in the definition of the equivalence class $[(a,b)]_{p}$ for modular
$p$, which is elucidated as follows: 
\[
\ensuremath{[(a,b)]_{p}:=\{(x,y)\in\mathbb{N}^{2}|\ x\equiv a\,(\textrm{mod}\,p),\ y\equiv b(\textrm{mod}\,p)\}}.
\]
 The ensemble of these equivalence classes is denoted as 
\[
\mathbf{Z}_{p}^{2}=\mathbf{Z}_{p}\times\mathbf{Z}_{p}=\{[(a,b)]_{p}|\ (a,b)\in\mathbb{N}^{2}\},
\]
where $\mathbf{Z}_{p}=\mathbb{Z}/p\mathbb{Z}$ on non-negative integers.

To illustrate, when training a model on 3-digit addition, we observe
that the learned operation function $f_{op}:\mathbb{N}^{2}\to\mathbb{N}$
effectively translates to $f_{op}(a,b)=f_{op}([(a,b)]_{10^{3}})$, which will be stated in the result section.

In alternative training data scenarios, the definition of equivalence classes necessitates adaptation to accommodate specific contexts.

\section{Results\label{sec:Experiments}}

In this section, we present the key results and findings from our experiments. These include observations on the phenomenon of generalization exhibited by the models, the learned algebraic structure, as well as the probability and representation structures in the model's learning process.

\subsection{Generalization in OOD Domain}
\begin{figure}[htb]
\begin{centering}
\includegraphics[width=0.95\linewidth]{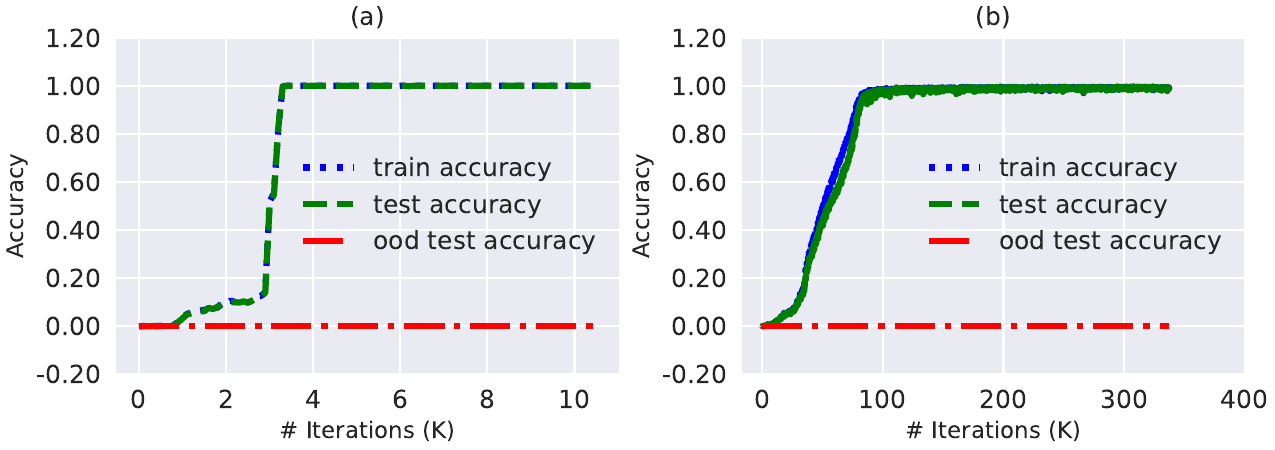}
\par\end{centering}
\caption{Training curves in addition and multiplication operations.}
\label{fig:training_curves}
\end{figure}

Figure~\ref{fig:training_curves} depicts the training, ID test, and OOD test accuracy for addition and multiplication operations in domains $\mathcal{D}_{1}, \mathcal{D}_{2}$, and $\mathcal{D}_{3}$ across different iterations. Panel (a) displays the training curve for addition learned by NanoGPT, while Panel (b) showcases the curve for multiplication learned by MinGPT. The hyperparameters employed by NanoGPT and MinGPT can be found in Table~\ref{table:hyperparameters}. 

By examining the figure, it becomes evident that both addition and multiplication quickly converge to a stable state, achieving (almost) 100\% accuracy in training and ID testing in $\mathcal{D}_{1}$ and $\mathcal{D}_{2}$. However, throughout the entire training process, the OOD test accuracy remains zero for both 3-digit addition and multiplication in $\mathcal{D}_{3}$. These results align with the discoveries made by \citet{jelassi2023length} and \citet{lee2023teaching}. When training on $n$-digit operations with $n$-digit operands, the models demonstrate excellent generalization on unseen $n$-digit inputs. Yet, they perform abysmally and mysteriously on longer, unseen cases, establishing a contrast between ID generalization and OOD generalization. Given the strikingly poor OOD performance, it is natural to question whether it solely stems from random errors or if there is any meaningful knowledge learned. The solution to the problem will be presented in the subsequent subsection.

\subsection{Algebraic Structure}

The mysterious absence of generalizability in the OOD domain prompts us to delve deeper into the results. We begin by examining some samples from domains $\mathcal{D}_{2}$ and $\mathcal{D}_{3}$. These examples are illustrated in Table~\ref{table:sample_results}. When observing the 3-digit addition and multiplication cases, we notice that the trained models produce incorrect results for the 4-digit instances. Strikingly, these erroneous outputs mirror the results obtained from the 3-digit cases. It appears that the model's outputs peculiarly ``disregard" the thousands digit of the input numbers, irrespective of whether it is an addition or multiplication operation.

\begin{table}[htb]
\small
\begin{centering}
\begin{tabular}{>{\centering}m{2.5cm}>{\centering}m{2.1cm}>{\centering}m{2.1cm}}
\toprule 
\centering{}\textbf{Operands} & \centering{}\textbf{Output Result} & \textbf{Correct Result}\tabularnewline
\midrule
349 + 705 & 1,054 & 1,054\tabularnewline
1,349 + 2,705 & 1,054 & 4,054\tabularnewline
128 $\times$ 256   & 32,768 & 32,768\tabularnewline
3,128 $\times$ 4,256   & 32,768 & 13,312,768\tabularnewline
\bottomrule
\end{tabular}
\par\end{centering}
\caption{Examples on models' outputs for addition and multiplication.}
\label{table:sample_results}
\end{table}
\begin{figure}[htb]
\begin{centering}
\includegraphics[width=0.99\linewidth]{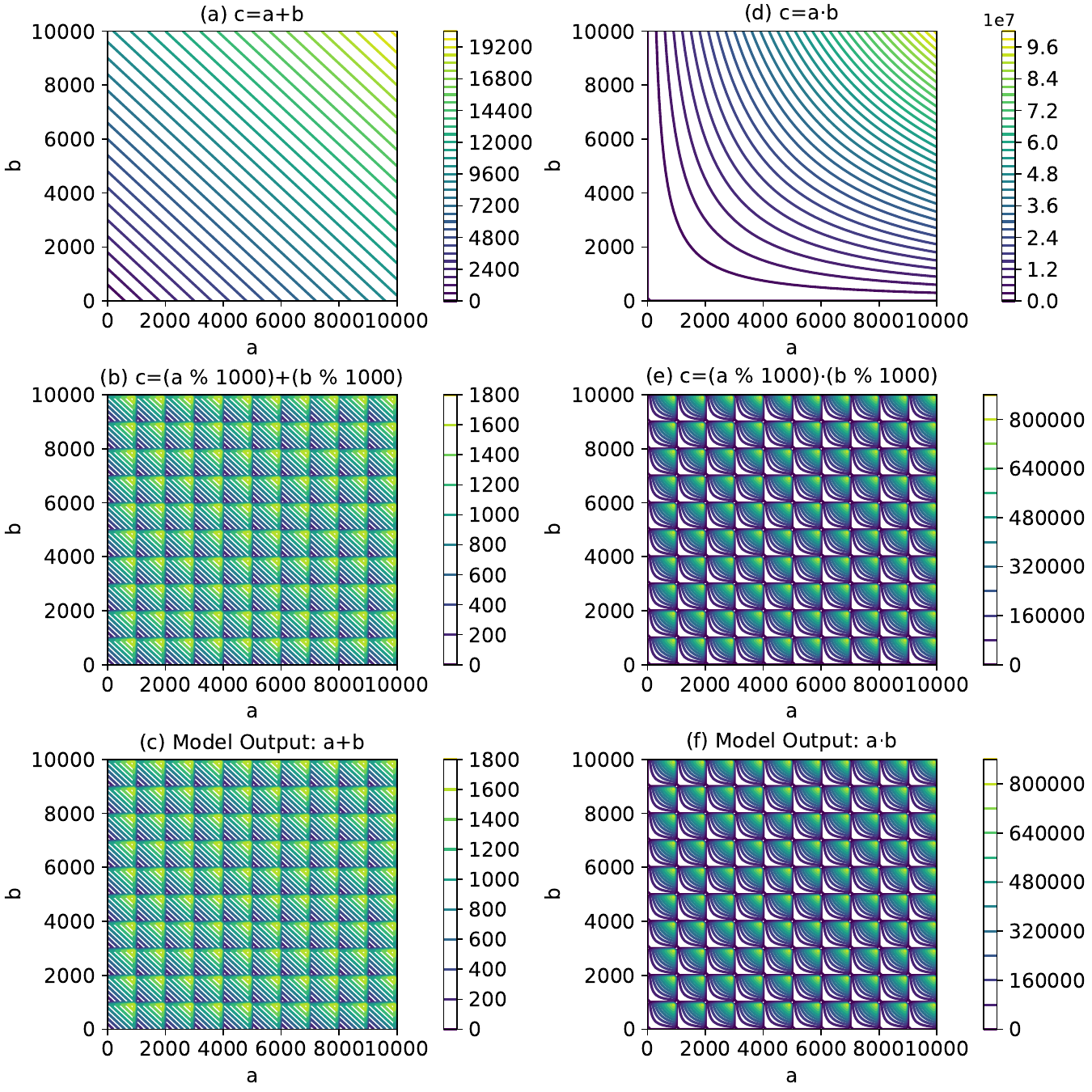}
\par\end{centering}
\caption{Contour plots for addition and multiplication operations. }
\label{fig:contour_plots}
\end{figure}
\begin{figure*}[tb]
\begin{centering}
\includegraphics[width=0.75\linewidth]{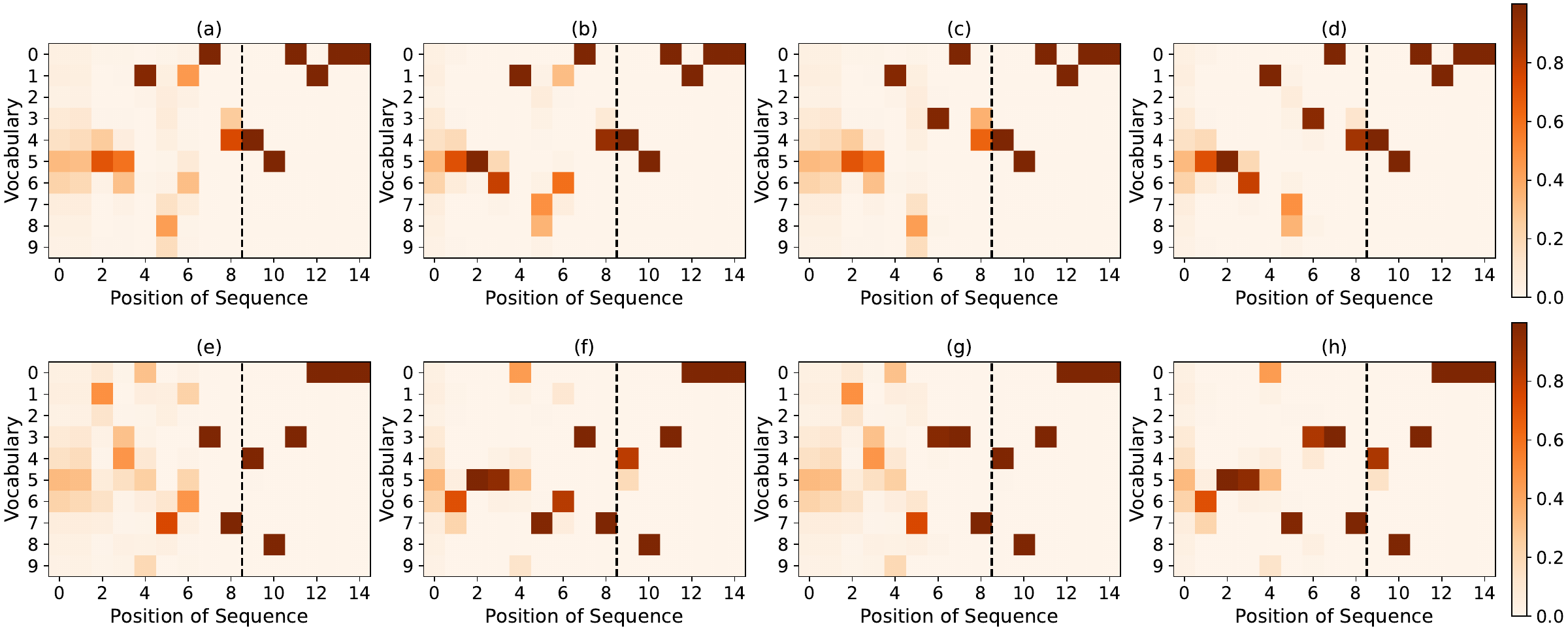}
\par\end{centering}
\caption{
The probability distribution of each digit of the sequence in an addition operation $c=a+b$. The left side of the black dashed line represents the input $a+b$, while the right side is the result $c$. Figure \ref{fig:addition_probs}(a) and Figure \ref{fig:addition_probs}(e) represent the $349+705$ and $128+256$, and the outputs are $1,054$  and $384$ ($450100$ and $483000$ in actual sequence output), respectively. In the second column, we perturb the thousands digit of $a$: Figure \ref{fig:addition_probs}(b) represents $1,349+705$, and Figure \ref{fig:addition_probs}(f) represents $3,128+256$. In the third column, we perturb the thousands digit of $b$: Figure \ref{fig:addition_probs}(c) represents $349+2,705$, and Figure \ref{fig:addition_probs}(g) represents $128+4,256$. In the fourth column, we simultaneously perturb the thousands digit of $a$ and $b$: Figure \ref{fig:addition_probs}(d) represents $1,349+2,705$, and Figure \ref{fig:addition_probs}(h) represents $3,128+4,256$.
}
\label{fig:addition_probs}
\end{figure*}


To systematically analyze the behavior in the OOD domain $\mathcal{D}_{3}$, we explore the entire two-dimensional lattice of 4-digit integers, namely $\mathbb{N}^2\cap [0,10^4)^2$. Figure~\ref{fig:contour_plots} presents the contour plots for the ground truth results of addition operation $c=a+b$ (Panel (a)) and multiplication operation $c=a\cdot b$ (Panel (d)), with the number $a$ on the horizontal axis and the number $b$ on the vertical axis. These landscapes represent the expected learning and generalization capabilities of the models on this lattice space.

However, when we utilize our trained models to generate results based on 3-digit operations, an unmistakably distinct pattern emerges, as depicted in Panel (c) for addition and Panel (f) for multiplication. This prompts us to investigate what structure the models have learned. We discover that there is a modular relationship between the operands $a$ and $b$. The learned structure can be represented as $c = (a \,\,\textrm{mod}\,\, 10^3) \circ (b \,\,\textrm{mod}\,\, 10^3)$, where $\circ$ represents either addition $+$ or multiplication $\times$. The ground truth landscapes of these functions on the 4-digit integer lattice are exhibited in Panel (b) for addition and Panel (e) for multiplication. Visually, these two panels are identical to Panel (c) and Panel (f), respectively. 
We compare the results of the operation $(a \,\,\textrm{mod}\,\, 10^3) \circ (b \,\,\textrm{mod}\,\, 10^3)$ with the outputs produced by the model. Surprisingly, they are identical across the entire space $\mathbb{N}^2\cap [0,10^4)^2$.

To formalize the results, we recall the definition of equivalence classes $[(a,b)]_{p}$ for modular $p=10^3$: 
\[
\ensuremath{[(a,b)]_{p}:=\{(x,y)\in\mathbb{N}^{2}|\ x\equiv a\,(\textrm{mod}\,p),\ y\equiv b(\textrm{mod}\,p)\}}.
\]
As $[(a,b)]_{p}$ is an equivalence class, we use the element in $\mathbb{N}^2\cap [0,10^3)^2=\mathcal{D}_1\cup\mathcal{D}_2$ to serve as the representative of the class. The ensemble of these equivalence classes then forms the space
\[
\mathbf{Z}_{p}^{2}=\mathbf{Z}_{p}\times\mathbf{Z}_{p}=\{[(a,b)]_{p}|\ (a,b)\in\mathbb{N}^{2}\},
\]
where $\mathbf{Z}_{p}=\mathbb{Z}/p\mathbb{Z}=\{[0]_p,[1]_p,\cdots,[p-1]_p\}$.

The models trained on 3-digit addition and multiplication actually learned the operation functions $f_{op}:\mathbf{Z}_p\times \mathbf{Z}_p\to\mathbb{N}$ for all integer paris on $\mathbb{N}\times \mathbb{N}$ such that $f_{op}(a,b)=f_{op}([(a,b)]_{p})$ with $p=10^{3}$. As an example, $f_{op}(1349,2705)=f_{op}([(349,705)]_{10^3})$. For addition, the learned operation is $f_{+}(1349,2705)=f_{+}([(349,705)]_{10^3})=1054$, while the learned multiplication is $f_{\times}(1349,2705)=f_{\times}([(349,705)]_{10^3})=246045$. 

As a summary of the results, the models learn to generalize the input in the OOD domain $\mathcal{D}_3$ by assimilating equivalence classes in the ID domain $\mathcal{D}_1\cup\mathcal{D}_2$. This result shows the limitations of the models. However, this capability allows the models to extend their learned knowledge beyond the ID domain $\mathcal{D}_1\cup\mathcal{D}_2$ shaped by the specific training data $\mathcal{D}_1$. Even though the output is wrong, it is not so bad. They have still managed to acquire useful information and demonstrate some level of learning. 

In order to gain a deeper understanding of the training process for Transformer models, we examine their token-level mapping using addition as an example. Consider two ($n$+1)-digit numbers, where $a=a_n\times 10^n+\cdots a_1\times 10+a_0$ and $b=b_n\times 10^n+\cdots b_1\times 10+b_0$. When training a Transformer model using randomly sampled ($n$+1)-digit numbers, the model learns an approximate mapping from the token-level input to the true function $c=a+b=c_{n+1}\times 10^{n+1}+\cdots c_1\times 10+c_0$. The learned approximation allows the model to perform classification for each digit of the resulting sum $c$, as follows:
$$f_{Trans}(a_n,\cdots,a_0,b_{n},\cdots,b_0)\approx (c_0,c_1,\cdots,c_{n+1}).$$
However, if the highest digit is completely absent from the training data and is instead padded with zeros, the training only guarantees learning of low $n$-digit addition. In other words:
$$f_{Trans}(0,a_{n-1},\cdots,a_0,0,b_{n-1},\cdots,b_0)\approx (c_0,c_1,\cdots,c_{n},0).$$
This limitation may explain why it is challenging to generalize to higher digits when the model is trained solely on lower digits. The absence of examples with higher digits restricts the model's ability to accurately predict and generalize beyond the low-digit addition it has been trained on. 

Building upon the observed algebraic structure discussed in the previous context of this subsection, we also know that when testing the Transformer models on higher digits that are non-zero, they do not significantly impact the classification of each digit of $c$. 

As a remark, it is important to note that when dealing with alternative training data scenarios, the definition of equivalence classes may need to be adjusted accordingly. For example, if the training data consists exclusively of 1 and 3-digit operations, while OOD testing involves 2 and 4-digit numbers, or if the training data includes 3-digit numbers for operand $a$ and 4-digit numbers for operand $b$, the equivalence classes would require redefining to account for these specific contexts.

\subsection{Probability Structure}
\begin{figure*}[htb]
\begin{centering}
\includegraphics[width=0.75\linewidth]{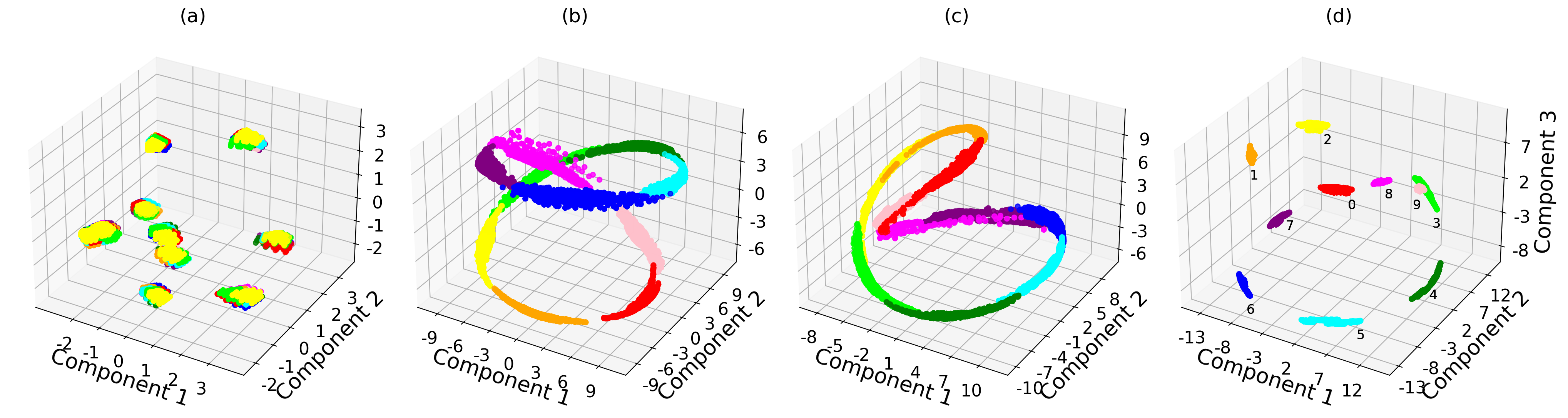}
\par\end{centering}
\caption{3D representation structure of the first three principal components in the addition operation. Figure \ref{fig:addition_representation_pca_3d}(a) to Figure \ref{fig:addition_representation_pca_3d}(d) represent the initial model, model with 14\%, 51\%, and 100\% test accuracy, respectively.}
\label{fig:addition_representation_pca_3d}
\end{figure*}

In the preceding subsection, we examined the structured algebraic patterns present in the output results. Considering that generative Transformer models generate outputs based on probability distributions, our model employs a greedy approach to select the output sequence with the maximum probability. We now shift our focus from algebraic structures to the probability distribution of the output sequences. Our objective is to investigate the underlying factors that contribute to the emergence of these structured algebraic patterns.

Specifically, we take two examples of 3-digit addition, namely $a + b$. We introduce perturbations to the thousands digit of both $a$ and $b$, enabling us to compare the variations in probability distributions before and after the perturbations occur. This comparative study will shed light on the mechanisms underlying the observed structured algebraic patterns.

Figure \ref{fig:addition_probs} displays the probability distributions of the next tokens in the vocabulary at each position within the sequence for two examples. The plot showcases the probabilities before and after perturbations for each token. Remarkably, we observe that regardless of whether we perturb $a$ and $b$ separately or simultaneously, the probability distribution in the model's output sequence remains largely unchanged.

Furthermore, we note that the digits with the highest probability in the output sequence remain consistent. This result implies that the algebraic structure of the model expands from $\mathbf{Z}_{p}\times \mathbf{Z}_{p}$ to $\mathbb{N}\times \mathbb{N}$. This expansion elucidates the structured patterns depicted in Figure \ref{fig:contour_plots}. Additionally, we conducted a systematic examination of the entire integer lattice within $\mathbb{N}^2\cap [0,10^4)^2$. Notably, the results obtained from this comprehensive analysis exhibit robustness, further supporting our findings.


\subsection{Representation Structure}
Within the probability structure, we made a significant observation that the model's output remains insensitive to perturbations in the thousands digit. This probability structure is rooted in the representation of the input sequence, which can be expressed as follows: $\mathbf{P} = \mathrm{Softmax}(\mathbf{WX}),$ where $\mathbf{P} \in [0,1]^{V\times L_{input}}$ represents the probability matrix for the next tokens at each position, $\mathbf{W} \in \mathbb{R}^{V\times d_{model}}$ signifies the learned weight matrix, and $\mathbf{X} \in \mathbb{R}^{d_{model} \times L_{input}}$ denotes the learned representation matrix of the input. The variables $V$, $L_{input}$, and $d_{model}$ correspond to the vocabulary size, input length, and model embedding dimension, respectively.

In this subsection, we delve deeper into the influence of these representations on the probability structure, thereby shedding light on their role in shaping the observed algebraic properties.

In order to explore the representations of $a+b$ in a systematic manner, we conducted a thorough analysis on the two-dimensional integer lattice of 4-digit numbers. Specifically, for each input sequence $a+b$, we obtained a high-dimensional embedding by considering the last column of the learned representation matrix $\mathbf{X}$. Subsequently, we applied principle component analysis (PCA) to project these embeddings into three dimensions.

Figure \ref{fig:addition_representation_pca_3d} showcases the four different phases of representation observed during the learning process of the model. The visualizations in the figure depict the representations using the first three principle components. More specifically, Figure \ref{fig:addition_representation_pca_3d}(a) to \ref{fig:addition_representation_pca_3d}(d) correspond to the random initial model, the model with approximately 14\%, 51\%, and 100\% ID test accuracy, respectively. The colors in each figure correspond to the true units digit of the resulting $a+b$. 

The observations made from Figure \ref{fig:addition_representation_pca_3d} demonstrate that the representations gradually transition from disorderly to structured throughout the learning process. Initially, the representations appear random with colors mixed together (Figure \ref{fig:addition_representation_pca_3d}(a)). However, as the training progresses, the structure of the learned representations becomes increasingly refined (Figure \ref{fig:addition_representation_pca_3d}(b) and (c)), ultimately leading to the development of a well-learned representation (Figure \ref{fig:addition_representation_pca_3d}(d)) where each color is separated according to its true label.

\subsection{From Representation to Algebraic Structure}

The findings discussed above regarding algebraic structures, probability distributions, and representation structures also hold true for multiplication operations. 

The systematic analysis approach outlined earlier provides a comprehensive understanding of the model's generalization capabilities through the assimilation of equivalence classes present within the ID domain. The representation structures successfully incorporate the assimilation of these equivalence classes, thereby extending the ID structure to OOD scenarios via the probability distribution of sequences. Consequently, this assimilation becomes evident within the algebraic structures as well.

\section{Robustness Studies}


In this section, we conduct thorough empirical analyses using various model sizes (GPT-Nano, GPT-Micro, GPT-Mini) and training data volumes, also exploring different datasets and encoding methods, to validate the robustness of our findings.

(1) \textbf{Encoding method:} For the main experiments, we chose the reversed encoding method for $n$-digit addition and multiplication, due to its faster convergence speed. Here we  test the alternative non-reversed encoding method and obtain consistent results (see $V_3$ in Table~\ref{table:variations}). The convergence time, consequently, is approximately 7.44 times longer than that of the reversed encoding method.

(2) \textbf{Scope of the dataset and training scheme:}
Additional experiments with variations in the training set include setting the rightmost digit to 0 (see $V_1$ in Table~\ref{table:variations}), setting the tens digit to 0 ($V_2$), and extending the OOD test to $10^6$ and $10^7$ ($V_4$). All the variations achieve 100\% accuracy for ID domain and 0\% for OOD domain. The results from $V_3$ and $V_4$ in OOD completely correspond with those of the equivalence class $[(a, b)]_{1000}$. Similarly, $V_1$ and $V_2$'s OOD results are totally consistent with these from the equivalence class $[(a, b)]_{10}$, as defined in the following equations ~(\ref{equ:v1}) and~(\ref{equ:v2}), respectively:

{\small
\begin{equation}
\label{equ:v1}
[(a, b)]_p:=\{(x, y) \in \mathbb{N}^2 \mid x \equiv \lfloor \tfrac{a}{p} \rfloor \cdot p, \ y \equiv \lfloor \tfrac{b}{p} \rfloor \cdot p \}.
\end{equation}
}
{\small
\begin{equation}
\label{equ:v2}
\begin{split}
[(a, b)]_p := \{ (x, y) \in \mathbb{N}^2 \mid & x \equiv \lfloor \tfrac{a}{10p} \rfloor \cdot 10p + a \bmod p, \\
& y \equiv \lfloor \tfrac{b}{10p} \rfloor \cdot 10p + b \bmod p \}.
\end{split}
\end{equation}
}

\begin{table}[htb]
\footnotesize
\begin{centering}
\begin{tabular}{>{\raggedright}m{3.3cm}>{\centering}m{1.0cm}>{\centering}m{1.0cm}}
\toprule
\textbf{Versions} & \textbf{ID} & \textbf{OOD}\tabularnewline
\midrule
$V_1$: rightmost digit be 0 & 100\% & 0\tabularnewline
$V_2$: tens digit be 0 & 100\% & 0\tabularnewline
$V_3$: non-reverse encoding & 100\% & 0 \tabularnewline
$V_4$: extended OOD & 100\% & 0 \tabularnewline
\bottomrule
\end{tabular}
\par\end{centering}
\caption{The accuracy of ID test and OOD test in different addition variations.}
\label{table:variations}
\end{table}

\begin{figure}[htb]
\begin{centering}
\includegraphics[width=0.75\linewidth]{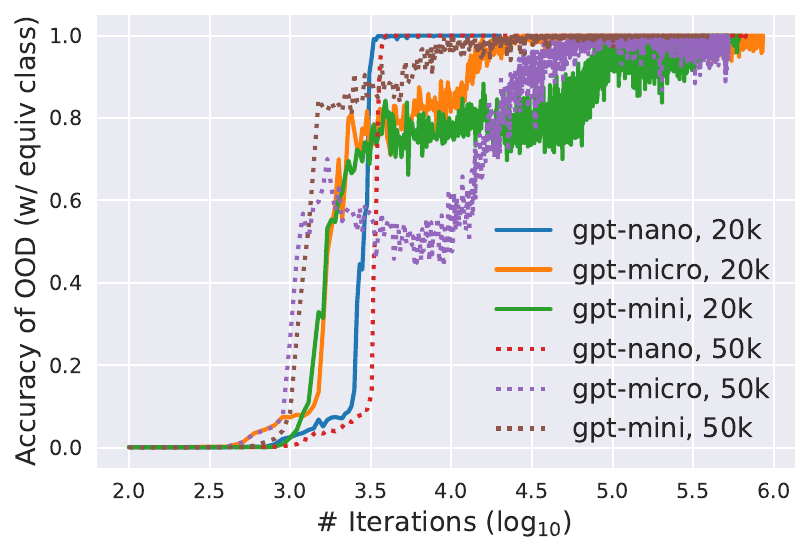}
\par\end{centering}
\caption{The accuracy of OOD test on equivalence for different model and data scales}
\label{fig:ood_equiv_class_acc}
\end{figure}

(3) \textbf{Model and data scales:}
To explore the potential applicability of our findings to large models, we conducted a detailed examination of outcomes across different model and data scales. Our analysis included three distinct model scales with increasing size: GPT-Nano, GPT-Micro, and GPT-Mini, as defined in the code. In addition to model size, we also evaluated the influence of varying training data sizes, specifically 20k and 50k, on the task of 3-digit addition. We focused on the accuracy of OOD test samples by comparing the model outputs with the results on $(a\%1000)+(b\%1000)$. As depicted in Figure \ref{fig:ood_equiv_class_acc}, there's a noticeable trend where, with progressing training, the above accuracy in all experiments approaches 100\%, and the algebraic structure of equivalence classes becomes more evident in OOD tests across different model and data scales. Notably, even as the model scales increase, our findings remain consistent. This consistency reinforces our confidence that these results might extend to larger language models (LLMs).

\section{Discussion}

In this section, we  discuss some aspects of our investigation. 
Our work corroborates the findings of \cite{anil2022exploring} that increasing the size of a model does not increase its ability to generalize across tasks of different lengths. Through careful robustness studies across a range of model sizes, we make similar observations with respect to equivalence generalization. This suggests that the ability of length generalization may be independent of model size, and that such findings may be applicable to large language models.

Another point to consider is that while we discovered equivalence classes defined by modular $(a\% p, b\% p)$, the output itself $(a\% p)+(b\% p)$ is not modular arithmetic $(a+b)\% p$. This is different from the direct study of modular arithmetic such as conducted in \citet{jelassi2023length}. Moreover, \citet{jelassi2023length} merely raises the question of why modulo 100 works effectively while modulo 101 fails, without exploring beyond the observation. In contrast, our study highlights the consistency between the definition of equivalence classes and modular arithmetic, enabling us to explain observed differences and offer insights into the behavior of the models.

\section{Conclusion\label{sec:Conclusion}}

We investigate the length generalization problem in arithmetic tasks for generative language models. We perform mechanistic analysis on smaller models and reveal that these models have strong generalization within the trained distribution. However, our investigation also uncovers an underlying algebraic structure that contributes to the models' unsatisfactory performance on OOD inputs. The models attempt to map OOD inputs using equivalence relations within the ID domain (we call \textit{``equivalence generalization"}), leading to errors and a lack of robustness in OOD scenarios. The representation plays a crucial role in enabling both ID and OOD generalization. The observation that length generalization ability does not vary with model scale, helps us extend our conclusion to LLMs.

Despite challenges in OOD generalization, our findings suggest that these models hold valuable information for improved generalization. However, due to the inherent subjectivity of natural language, much more efforts are needed to establish equivalence in NLP tasks for LLMs. In addition, the finding of equivalence generalization may serve as helpful prior knowledge, guiding the training process of LLMs regarding generalizability. For example, we may stop training once these equivalence classes are formed, reducing the extensive data needed for generalizability. Besides, in domain adaptation, people often finetune existing models, to adapt to OOD data and similarity metrics of equivalence classes may facilitate this process.

\section*{Acknowledgments}
This work is supported by Shanghai Artificial Intelligence Laboratory.

\section*{Contribution Statement}
Xingcheng Xu and Zihao Pan contributed equally in this work. 

\bibliographystyle{named}
\bibliography{reference_ood}

\begin{thebibliography}{}

\bibitem[\protect\citeauthoryear{Akhtar and Ragavendran}{2020}]{akhtar2020interpretation}
Nadeem Akhtar and U~Ragavendran.
\newblock Interpretation of intelligence in cnn-pooling processes: a methodological survey.
\newblock {\em Neural computing and applications}, 32(3):879--898, 2020.

\bibitem[\protect\citeauthoryear{Anil \bgroup \em et al.\egroup }{2022}]{anil2022exploring}
Cem Anil, Yuhuai Wu, Anders Andreassen, Aitor Lewkowycz, Vedant Misra, Vinay Ramasesh, Ambrose Slone, Guy Gur-Ari, Ethan Dyer, and Behnam Neyshabur.
\newblock Exploring length generalization in large language models.
\newblock {\em Advances in Neural Information Processing Systems}, 35:38546--38556, 2022.

\bibitem[\protect\citeauthoryear{Anthropic}{2023}]{Anthropic2023Claude}
Anthropic.
\newblock Model card and evaluations for claude models.
\newblock 2023.

\bibitem[\protect\citeauthoryear{Barbiero \bgroup \em et al.\egroup }{2022}]{barbiero2022entropy}
Pietro Barbiero, Gabriele Ciravegna, Francesco Giannini, Pietro Li{\'o}, Marco Gori, and Stefano Melacci.
\newblock Entropy-based logic explanations of neural networks.
\newblock In {\em Proceedings of the AAAI Conference on Artificial Intelligence}, volume~36, pages 6046--6054, 2022.

\bibitem[\protect\citeauthoryear{Bender \bgroup \em et al.\egroup }{2021}]{bender2021dangers}
Emily~M Bender, Timnit Gebru, Angelina McMillan-Major, and Shmargaret Shmitchell.
\newblock On the dangers of stochastic parrots: Can language models be too big?
\newblock In {\em Proceedings of the 2021 ACM conference on fairness, accountability, and transparency}, pages 610--623, 2021.

\bibitem[\protect\citeauthoryear{Brown \bgroup \em et al.\egroup }{2020}]{brown2020language}
Tom Brown, Benjamin Mann, Nick Ryder, Melanie Subbiah, Jared~D Kaplan, Prafulla Dhariwal, Arvind Neelakantan, Pranav Shyam, Girish Sastry, Amanda Askell, et~al.
\newblock Language models are few-shot learners.
\newblock {\em Advances in neural information processing systems}, 33:1877--1901, 2020.

\bibitem[\protect\citeauthoryear{Bubeck \bgroup \em et al.\egroup }{2023}]{bubeck2023sparks}
S{\'e}bastien Bubeck, Varun Chandrasekaran, Ronen Eldan, Johannes Gehrke, Eric Horvitz, Ece Kamar, Peter Lee, Yin~Tat Lee, Yuanzhi Li, Scott Lundberg, et~al.
\newblock Sparks of artificial general intelligence: Early experiments with gpt-4.
\newblock {\em arXiv preprint arXiv:2303.12712}, 2023.

\bibitem[\protect\citeauthoryear{Chowdhery \bgroup \em et al.\egroup }{2023}]{chowdhery2023palm}
Aakanksha Chowdhery, Sharan Narang, Jacob Devlin, Maarten Bosma, Gaurav Mishra, Adam Roberts, Paul Barham, Hyung~Won Chung, Charles Sutton, Sebastian Gehrmann, et~al.
\newblock Palm: Scaling language modeling with pathways.
\newblock {\em Journal of Machine Learning Research}, 24(240):1--113, 2023.

\bibitem[\protect\citeauthoryear{Dubois \bgroup \em et al.\egroup }{2019}]{dubois2019location}
Yann Dubois, Gautier Dagan, Dieuwke Hupkes, and Elia Bruni.
\newblock Location attention for extrapolation to longer sequences.
\newblock {\em arXiv preprint arXiv:1911.03872}, 2019.

\bibitem[\protect\citeauthoryear{Jelassi \bgroup \em et al.\egroup }{2023}]{jelassi2023length}
Samy Jelassi, St{\'e}phane d'Ascoli, Carles Domingo-Enrich, Yuhuai Wu, Yuanzhi Li, and Fran{\c{c}}ois Charton.
\newblock Length generalization in arithmetic transformers.
\newblock {\em arXiv preprint arXiv:2306.15400}, 2023.

\bibitem[\protect\citeauthoryear{Karpathy}{2022}]{karpathy2022mingpt}
Andrej Karpathy.
\newblock A minimal pytorch re-implementation of the openai gpt (generative pretrained transformer) training.
\newblock {\em GitHub https://github.com/karpathy/minGPT}, 2022.

\bibitem[\protect\citeauthoryear{Lee \bgroup \em et al.\egroup }{2023}]{lee2023teaching}
Nayoung Lee, Kartik Sreenivasan, Jason~D Lee, Kangwook Lee, and Dimitris Papailiopoulos.
\newblock Teaching arithmetic to small transformers.
\newblock {\em arXiv preprint arXiv:2307.03381}, 2023.

\bibitem[\protect\citeauthoryear{Liu \bgroup \em et al.\egroup }{2022}]{liu2022towards}
Ziming Liu, Ouail Kitouni, Niklas~S Nolte, Eric Michaud, Max Tegmark, and Mike Williams.
\newblock Towards understanding grokking: An effective theory of representation learning.
\newblock {\em Advances in Neural Information Processing Systems}, 35:34651--34663, 2022.

\bibitem[\protect\citeauthoryear{Nam \bgroup \em et al.\egroup }{2020}]{nam2020relative}
Woo-Jeoung Nam, Shir Gur, Jaesik Choi, Lior Wolf, and Seong-Whan Lee.
\newblock Relative attributing propagation: Interpreting the comparative contributions of individual units in deep neural networks.
\newblock In {\em Proceedings of the AAAI conference on artificial intelligence}, volume~34, pages 2501--2508, 2020.

\bibitem[\protect\citeauthoryear{Nanda and Lieberum}{2022}]{nanda2022mechanistic}
Neel Nanda and Tom Lieberum.
\newblock A mechanistic interpretability analysis of grokking.
\newblock In {\em Alignment Forum}, 2022.

\bibitem[\protect\citeauthoryear{Nogueira \bgroup \em et al.\egroup }{2021}]{nogueira2021investigating}
Rodrigo Nogueira, Zhiying Jiang, and Jimmy Lin.
\newblock Investigating the limitations of transformers with simple arithmetic tasks.
\newblock {\em arXiv preprint arXiv:2102.13019}, 2021.

\bibitem[\protect\citeauthoryear{OpenAI}{2023}]{OpenAI2023GPT4TR}
OpenAI.
\newblock Gpt-4 technical report.
\newblock {\em ArXiv}, abs/2303.08774, 2023.

\bibitem[\protect\citeauthoryear{Ouyang \bgroup \em et al.\egroup }{2022}]{ouyang2022training}
Long Ouyang, Jeffrey Wu, Xu~Jiang, Diogo Almeida, Carroll Wainwright, Pamela Mishkin, Chong Zhang, Sandhini Agarwal, Katarina Slama, Alex Ray, et~al.
\newblock Training language models to follow instructions with human feedback.
\newblock {\em Advances in Neural Information Processing Systems}, 35:27730--27744, 2022.

\bibitem[\protect\citeauthoryear{Qian \bgroup \em et al.\egroup }{2022}]{qian2022limitations}
Jing Qian, Hong Wang, Zekun Li, Shiyang Li, and Xifeng Yan.
\newblock Limitations of language models in arithmetic and symbolic induction.
\newblock {\em arXiv preprint arXiv:2208.05051}, 2022.

\bibitem[\protect\citeauthoryear{Touvron \bgroup \em et al.\egroup }{2023a}]{touvron2023llama}
Hugo Touvron, Thibaut Lavril, Gautier Izacard, Xavier Martinet, Marie-Anne Lachaux, Timoth{\'e}e Lacroix, Baptiste Rozi{\`e}re, Naman Goyal, Eric Hambro, Faisal Azhar, et~al.
\newblock Llama: Open and efficient foundation language models.
\newblock {\em arXiv preprint arXiv:2302.13971}, 2023.

\bibitem[\protect\citeauthoryear{Touvron \bgroup \em et al.\egroup }{2023b}]{touvron2023llama2}
Hugo Touvron, Louis Martin, Kevin Stone, Peter Albert, Amjad Almahairi, Yasmine Babaei, Nikolay Bashlykov, Soumya Batra, Prajjwal Bhargava, Shruti Bhosale, et~al.
\newblock Llama 2: Open foundation and fine-tuned chat models.
\newblock {\em arXiv preprint arXiv:2307.09288}, 2023.

\bibitem[\protect\citeauthoryear{Trummer}{2022}]{trummer2022codexdb}
Immanuel Trummer.
\newblock Codexdb: Synthesizing code for query processing from natural language instructions using gpt-3 codex.
\newblock {\em Proceedings of the VLDB Endowment}, 15(11):2921--2928, 2022.

\bibitem[\protect\citeauthoryear{Xuanyuan \bgroup \em et al.\egroup }{2023}]{xuanyuan2023global}
Han Xuanyuan, Pietro Barbiero, Dobrik Georgiev, Lucie~Charlotte Magister, and Pietro Li{\'o}.
\newblock Global concept-based interpretability for graph neural networks via neuron analysis.
\newblock In {\em Proceedings of the AAAI Conference on Artificial Intelligence}, volume~37, pages 10675--10683, 2023.

\bibitem[\protect\citeauthoryear{Yuan \bgroup \em et al.\egroup }{2019}]{yuan2019interpreting}
Hao Yuan, Yongjun Chen, Xia Hu, and Shuiwang Ji.
\newblock Interpreting deep models for text analysis via optimization and regularization methods.
\newblock In {\em Proceedings of the AAAI Conference on Artificial Intelligence}, volume~33, pages 5717--5724, 2019.

\bibitem[\protect\citeauthoryear{Yuan \bgroup \em et al.\egroup }{2020}]{yuan2020xgnn}
Hao Yuan, Jiliang Tang, Xia Hu, and Shuiwang Ji.
\newblock Xgnn: Towards model-level explanations of graph neural networks.
\newblock In {\em Proceedings of the 26th ACM SIGKDD International Conference on Knowledge Discovery \& Data Mining}, pages 430--438, 2020.

\bibitem[\protect\citeauthoryear{Zhong \bgroup \em et al.\egroup }{2023}]{zhong2023clock}
Ziqian Zhong, Ziming Liu, Max Tegmark, and Jacob Andreas.
\newblock The clock and the pizza: Two stories in mechanistic explanation of neural networks.
\newblock {\em arXiv preprint arXiv:2306.17844}, 2023.

\bibitem[\protect\citeauthoryear{Zong and Krishnamachari}{2023}]{zong2023solving}
Mingyu Zong and Bhaskar Krishnamachari.
\newblock Solving math word problems concerning systems of equations with gpt-3.
\newblock In {\em Proceedings of the AAAI Conference on Artificial Intelligence}, volume~37, pages 15972--15979, 2023.

\end{thebibliography}

\end{document}